\documentclass[journal]{IEEEtran}

\usepackage{amsmath}
\usepackage{booktabs}
\usepackage[justification=centering]{caption}
\usepackage[bottom=0.98in,top=0.75in,left=0.625in,right=0.625in]{geometry}

\ifCLASSINFOpdf
\else
\fi

\usepackage{xcolor}
\usepackage{graphicx}
\usepackage{comment}

\begin{document}

\title{LLMs meet Federated Learning for Scalable and Secure IoT Management}
%
%
%

\author{
    \IEEEauthorblockN{
    Yazan Otoum\IEEEauthorrefmark{1}, Arghavan Asad\IEEEauthorrefmark{1}, Amiya Nayak\IEEEauthorrefmark{2} \\} \vspace{1em}
  \IEEEauthorblockA{\small \IEEEauthorrefmark{1}School of Computer Science and Technology, Algoma University, Canada}
        \\ \IEEEauthorblockA{\small \IEEEauthorrefmark{2}School of Electrical Engineering and Computer Science, University  of Ottawa, Canada}

}

\maketitle

\thispagestyle{empty}
\pagestyle{empty}

\begin{abstract}
The rapid expansion of IoT ecosystems introduces severe challenges in scalability, security, and real-time decision-making. Traditional centralized architectures struggle with latency, privacy concerns, and excessive resource consumption, making them unsuitable for modern large-scale IoT deployments. This paper presents a novel Federated Learning-driven Large Language Model (FL-LLM) framework, designed to enhance IoT system intelligence while ensuring data privacy and computational efficiency. The framework integrates Generative IoT (GIoT) models with a Gradient Sensing Federated Strategy (GSFS), dynamically optimizing model updates based on real-time network conditions. By leveraging a hybrid edge-cloud processing architecture, our approach balances intelligence, scalability, and security in distributed IoT environments. Evaluations on the IoT-23 dataset demonstrate that our framework improves model accuracy, reduces response latency, and enhances energy efficiency, outperforming traditional FL techniques (i.e., FedAvg, FedOpt). These findings highlight the potential of integrating LLM-powered federated learning into large-scale IoT ecosystems, paving the way for more secure, scalable, and adaptive IoT management solutions.
\end{abstract}


%
\IEEEpeerreviewmaketitle

\section{Introduction}

\IEEEPARstart{T}{he} rapid proliferation of Internet of Things (IoT) ecosystems has led to a surge in connected devices, generating massive amounts of heterogeneous data. Managing and optimizing these complex systems presents significant challenges, particularly in real-time data processing, resource allocation, security, and energy efficiency \cite{said2023scalable}. Traditional cloud-centric approaches struggle with scalability, as they introduce latency, create potential single points of failure, and raise critical privacy concerns, especially in sensitive domains such as healthcare, smart cities, and industrial automation \cite{nazari2020integration, vo2022edge}. These limitations highlight the need for decentralized, intelligent, and adaptive frameworks capable of handling dynamic and resource-constrained IoT environments \cite{otoum2024advancing}. Recent advancements in artificial intelligence, particularly the emergence of LLMs, have demonstrated remarkable natural language understanding, decision-making, and predictive analytics capabilities. Unlike traditional machine learning models, which are task-specific and require extensive feature engineering, LLMs excel in contextual reasoning, multimodal learning, and adaptive decision-making. This makes them particularly valuable in IoT environments, where heterogeneous devices generate vast, unstructured data streams that need to be processed efficiently. LLMs can enable IoT systems to autonomously interpret vast amounts of data, optimize resource allocation, and enhance anomaly detection \cite{zong2025integrating}. However, deploying LLMs in IoT environments introduces several challenges, including computational constraints at the edge, real-time processing demands, and the need for privacy-preserving learning mechanisms. The direct application of LLMs in IoT systems without appropriate adaptations can lead to inefficiencies, high communication costs, and increased energy consumption \cite{tran2025energy}. Federated Learning (FL) has emerged as a promising paradigm for addressing these challenges by enabling collaborative model training across distributed devices without requiring raw data transfer. FL allows IoT devices to locally train models and share only aggregated updates, preserving data privacy while reducing communication overhead \cite{kairouz2021advances}. By integrating LLMs with FL in a hybrid edge-cloud architecture, we can leverage the strengths of both paradigms—LLMs for intelligent decision-making and FL for scalable, privacy-preserving learning. This integration facilitates adaptive learning strategies where models dynamically update based on real-time IoT data while minimizing latency and ensuring robust security. To bridge this gap, we propose an LLM-driven adaptive IoT management framework that incorporates federated learning for decentralized training and a hybrid edge-cloud architecture for optimal computational distribution. Our framework introduces a novel gradient sensing federated strategy (GSFS) that dynamically regulates client participation and model updates based on real-time performance metrics. This approach optimizes communication efficiency while maintaining high model accuracy and adaptability across diverse IoT environments. The key contributions of this paper are as follows:
\begin{itemize}
\item The proposed model introduces a GSFS that dynamically optimizes client participation, learning rates, and model aggregation based on efficient system performance. Our approach reduces communication overhead and improves model convergence compared to FedAvg and FedOpt.
\item We design and implement a hybrid edge-cloud processing architecture that enables low-latency, on-device intelligence while leveraging cloud resources for global model refinement. This approach ensures scalability and efficiency in heterogeneous IoT deployments.
\item We conduct an extensive evaluation using the IoT-23 dataset, demonstrating that GSFS achieves 0.63\% higher accuracy on the central model, 1.02\% on the client model, 21.57\% lower response latency on the central side and 51.45\% on the client side, and improved energy efficiency compared to FedAvg, also a similar trend with FedOpt, demonstrating its improvement in federated learning.
\end{itemize}
The remainder of this paper is structured as follows: Section~\ref{sec:Background} reviews related work on federated learning, LLMs, and their applications in IoT environments. Section~\ref{sec:ProposedFramework} introduces the proposed LLM-FL hybrid framework, detailing its key components and adaptive learning strategies. Section~\ref{sec:Results} presents the performance analysis, comparing our approach against baseline models. Finally, Section~\ref{sec:Conclusion} concludes the paper by summarizing our findings and outlining future research directions.

\begin{figure*}[!t]
    \centering
    \includegraphics[scale=0.45]{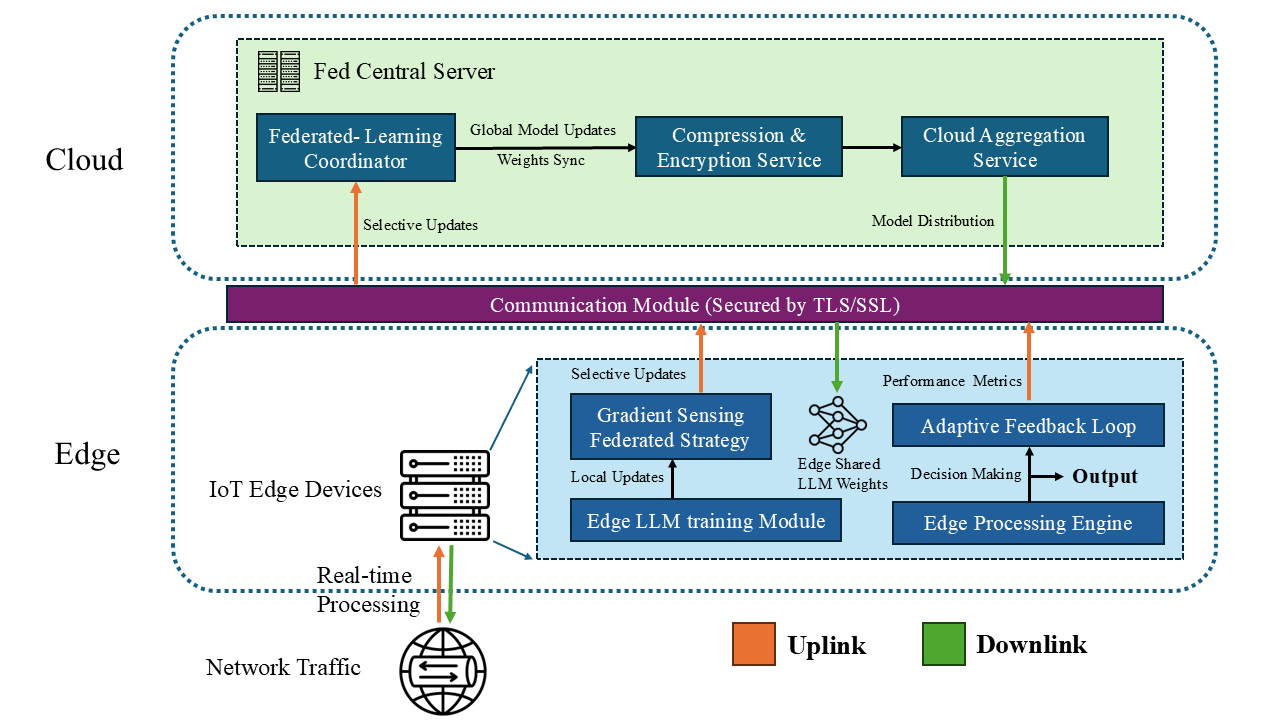}  
    \caption{Architecture of our proposed IoT management system integrating edge and cloud computing with federated learning. The framework includes key components such as local LLM training at the edge, a federated learning coordinator for global model updates, a communication module for secure and efficient data transfer, etc..}
    \label{fig:arch}
\end{figure*}

\section{Related Work}
\label{sec:Background}
Existing approaches to IoT management primarily rely on centralized data processing, which poses scalability and privacy concerns. Recent advancements in federated learning and edge computing have enabled distributed and privacy-preserving solutions. Open-source LLMs have also demonstrated the potential to improve data management and decision-making. However, the integration of these technologies within IoT systems remains underexplored. This study builds on prior work by proposing an integrated framework leveraging these advancements to optimize IoT operations. The authors of \cite{an2024iot} address the challenge of LLM outputs violating physical laws in IoT environments. By integrating LLMs with IoT systems, their framework ensures task reasoning aligns with real-world constraints. Their approach demonstrated an improvement in the accuracy of IoT task reasoning compared to conventional methods. The integration of physical law checks reduced anomalous outputs, enhancing trust in automated decision-making systems. The authors of \cite{wang2024flora} propose FLoRA, a method for federated fine-tuning of LLMs using heterogeneous low-rank adaptations. This method allows efficient personalization across diverse datasets without sharing raw data. Experiments showed that FLoRA achieved model performance within 1.5\% of centralized fine-tuning methods while reducing communication costs. This validated the effectiveness of FL for LLM adaptation in data-sensitive environments. The authors of \cite{ye2024openfedllm} present OpenFedLLM, a framework for training LLMs on decentralized private data using FL. This method demonstrates privacy-preserving training for LLMs while maintaining competitive performance. OpenFedLLM achieved comparable performance to centralized training with less than a 2\% accuracy gap on NLP benchmarks. The study also reported significant improvements in privacy metrics, making it suitable for sensitive domains like healthcare IoT.


\section{Proposed Framework}
\label{sec:ProposedFramework}

\subsection{Architecture Overview}

The proposed framework integrates edge computing, federated learning, and cloud-based coordination to create an adaptive and scalable IoT management system. The architecture consists of the following key components:

\textbf{1) Local LLM Training Module:} 
Deployed on edge devices, this module processes telemetry data, system logs, and real-time sensor information to enable privacy-preserving decision-making. It utilizes lightweight fine-tuning techniques and knowledge distillation to reduce computational overhead. By training on-device, the framework minimizes data transmission costs while improving response latency.

\textbf{2) Federated Learning Coordinator:} 
A cloud-based entity is responsible for aggregating model updates from edge devices while preserving data privacy. The coordinator employs secure aggregation techniques such as Differential Privacy (DP-FedAvg) and Gradient Compression to optimize bandwidth usage. Periodic synchronization ensures system-wide adaptability while mitigating the impact of non-IID (non-identically distributed) data across IoT nodes.

\textbf{3) Prompt Management Module:} 
This module dynamically adjusts prompts based on context, device constraints, and real-time system requirements to optimize LLM-driven responses for specific IoT tasks. The cloud refines these prompts for broader analytical tasks, while edge devices use optimized query structures for low-latency decision-making.

\textbf{4) Communication Module:} 
Ensures secure and efficient data transmission between edge devices and cloud infrastructure. This module employs adaptive data compression techniques to reduce bandwidth consumption and TLS/SSL encryption to ensure the integrity and confidentiality of transmitted updates.

\textbf{5) Adaptive Feedback Loop:} 
This module continuously monitors performance metrics and user feedback to refine both model updates and prompt engineering. By leveraging reinforcement learning-based optimization, the framework dynamically adapts federated learning rates and client participation to enhance overall efficiency.


\subsection{Implementation Details}

\textbf{Edge Processing:} Each edge device is equipped with a Local Inference and Adaptation Engine that leverages trained LLMs for real-time decision-making. This enables critical tasks such as dynamic resource allocation, predictive maintenance, and anomaly detection without relying on cloud resources. By performing computations locally, the framework minimizes latency, reduces bandwidth consumption, and enhances overall energy efficiency, making it well-suited for resource-constrained IoT environments.

\subsubsection{Classical Federated Learning Baselines}

\paragraph{FedAvg.}
Federated Averaging (FedAvg) is one of the most classical and widely used federated learning strategies. Each client $k$ maintains a local model $\theta_{k,t}$ and updates it by running $E$ local epochs of stochastic gradient descent (SGD) on its private dataset $D_k$ with learning rate $\eta$. Formally, after local training, each client has

\begin{equation}
\theta_{k,t+1} \;=\; \theta_{k,t} - \eta\,\nabla \mathcal{L}(\theta_{k,t}, D_k).
\label{eq:update_rule}
\end{equation}

repeated for $E$ epochs. All clients then send their updated parameters (or gradients) to the central server, which performs a global aggregation. If there are $N$ clients in total and client $k$'s dataset size is $|D_k|$, a common aggregation is the weighted average:

\begin{equation}
\theta_{\text{global}}^{(t+1)} 
\;=\;
\frac{1}{\sum_{k=1}^N |D_k|}
\sum_{k=1}^N |D_k|\; \theta_{k,t+1}.
\label{eq:global_update}
\end{equation}

Finally, the new global model is broadcast back to each client, completing one federated round. 
\paragraph{FedOpt.}
FedOpt is a family of methods that extend FedAvg by using an \emph{adaptive optimizer} on the server side. Instead of simple averaging, the server updates the global model $\theta_{\text{global}}^{(t)}$ using certain rules, here, for instance, an Adam-like rule:
\begin{equation}
g_t \;=\; \frac{1}{N}\,\sum_{k=1}^N \Delta \theta_{k,t}.
\label{eq:global_gradient}
\end{equation}

\begin{equation}
m_{t+1} = \beta_1 m_t + (1-\beta_1) g_t.
\label{eq:momentum_update}
\end{equation}

\begin{equation}
v_{t+1} = \beta_2 v_t + (1-\beta_2) g_t^2.
\label{eq:variance_update}
\end{equation}

\begin{equation}
\theta_{\text{global}}^{(t+1)} \;=\; \theta_{\text{global}}^{(t)} - \eta_{\text{server}} \,\frac{m_{t+1}}{\sqrt{v_{t+1}} + \epsilon}.
\label{eq:global_update_rule}
\end{equation}

where $m_t$ and $v_t$ are first and second moment estimates similar to Adam. 

\textbf{Gradient Sensing Federal Strategy}: Most classical federated learning algorithms (e.g., FedAvg) require each client to upload its complete update (either model parameters or gradients) to the central server after each local training round. The server then performs global aggregation (e.g., simple or weighted averaging) and broadcasts the updated global model to all clients. This frequent and uniform synchronization results in high communication overhead and increased latency, particularly when the number of participating devices is large or network bandwidth is constrained. The GSFS minimizes unnecessary communication by enabling clients to \emph{asynchronously trigger} uploads based on significant locally observed changes. The core idea is to monitor the magnitude of updates or performance shifts in local models and initiate uploads only when a predefined \emph{threshold} is exceeded. The next section outlines the details of this approach from both the client and server perspectives.



\subsubsection{Client Side}

\paragraph{Performance-Based Trigger.}
Each client $k$ maintains a local model $\theta_{k,t}$ and a local validation dataset $D_{k}^{\text{val}}$. After several local mini-batch updates or at periodic intervals, the client evaluates its model on $D_{k}^{\text{val}}$ to obtain a performance metric (e.g., accuracy, loss, F1-score). Let $\mathcal{M}(\theta_{k,t}; D_{k}^{\text{val}})$ denote the chosen metric for client $k$ at time $t$. The client compares the current performance $\mathcal{M}(\theta_{k,t}; D_{k}^{\text{val}})$ with the performance at the last upload reference point $\mathcal{M}(\theta_{k,t_{\text{ref}}}; D_{k}^{\text{val}})$:
\begin{equation}\label{eq:perf-change}
\Delta \mathcal{M}_{k} \;=\; 
\Bigl|\; \mathcal{M}(\theta_{k,t}; D_{k}^{\text{val}})\;-\;\mathcal{M}(\theta_{k,t_{\text{ref}}}; D_{k}^{\text{val}})\Bigr|.
\end{equation}
If $\Delta \mathcal{M}_{k}$ exceeds an adaptive threshold $\delta_k^{\text{perf}}$, the client initiates an \emph{asynchronous upload} of its gradients updates to the server:\newpage
\begin{equation}\label{eq:perf-trigger}
\text{Trigger:} \quad 
\Delta \mathcal{M}_{k} \;>\;\delta_k^{\text{perf}}.
\end{equation}

\paragraph{Adaptive Threshold Based on Gradient Norms.}
In addition to using performance metrics, clients can monitor the norm of local gradients or parameter changes across training steps. Denote by $g_{k}^{(l)}$ the aggregated gradient norm at layer $l$ of client $k$. A moving average (mean) $\mu_{k}^{(l)}$ and standard deviation $\sigma_{k}^{(l)}$ can be maintained over recent training steps:
\begin{align}
\mu_{k}^{(l)} &\leftarrow \alpha\,\mu_{k}^{(l)} \;+\; (1-\alpha)\,\|\,g_{k}^{(l)}\|, \\
\sigma_{k}^{(l)} &\leftarrow \alpha\,\sigma_{k}^{(l)} \;+\; (1-\alpha)\,\Bigl\lvert\,\|\,g_{k}^{(l)}\|\;-\;\mu_{k}^{(l)}\Bigr\rvert,
\end{align}
where $\alpha \in (0,1)$ is a smoothing factor. An adaptive threshold $\delta_{k}^{(l)}$ can then be defined for each layer:
\begin{equation}\label{eq:adaptive-threshold}
\delta_{k}^{(l)} \;=\; \mu_{k}^{(l)} \;+\; \beta\,\sigma_{k}^{(l)},
\end{equation}
where $\beta$ is a hyperparameter that controls sensitivity. If any layer's gradient norm $\|\,g_{k}^{(l)}\|$ exceeds $\delta_{k}^{(l)}$, client $k$ triggers an upload. 

\subsubsection{Server Side}

\paragraph{Caching and Aggregation Trigger.}
When the central server receives a gradient update or parameter update $\Delta \theta_{k,t}$ from client $k$, it stores it in an \emph{update pool} along with metadata such as timestamp or training round:
\begin{equation}
\text{Pool} \;=\; \Bigl\{\,(\Delta \theta_{k,t},\, \text{time}_{k,t},\, k) \Bigr\}.
\end{equation}

Once the number of distinct clients $|\text{Pool}|$ reaches threshold $M$ (which we default to 60\% of all clients), the server performs a global aggregation step. Denoting by $\mathcal{K}_t$ the set of clients whose updates are included at aggregation time $t$, the form of the aggregation can be:

\begin{equation}\label{eq:fed-agg}
\theta_{\text{global}}^{(t+1)} 
\;=\; 
\text{Aggregate}\Bigl(\theta_{\text{global}}^{(t)},\,\{\Delta \theta_{k,t} : k \in \mathcal{K}_t\}\Bigr).
\end{equation}
We adopt a common choice to the weighted sum of the updates while using the submission frequency as the weights for calculation, for instance:
\begin{equation}\label{eq:weighted-fedavg}
\theta_{\text{global}}^{(t+1)}
\;=\; 
\theta_{\text{global}}^{(t)} 
\;+\;
\sum_{k \in \mathcal{K}_t} \alpha_k \,\Delta \theta_{k,t},
\end{equation}

where $\alpha_k$ reflects the importance of client $k$ (e.g., positively proportional to $k$'s submission frequency).

\subsubsection{Upload and Broadcast Design}

\paragraph{Consistent Model Broadcasting.}

After the server aggregates the updates in the pool to form a new global model $\theta_{\text{global}}^{(t+1)}$, it broadcasts this model to \emph{all} clients simultaneously:
\begin{equation}
\text{Broadcast:} \quad
\forall k,\quad \theta_{k}^{(t+1)} \leftarrow \theta_{\text{global}}^{(t+1)}.
\end{equation}\\
Each client receives the latest version of the global model, ensuring \emph{download consistency}. Clients will then reset their local reference models (e.g., $\theta_{k,t_{\text{ref}}} \leftarrow \theta_{k}^{(t+1)}$) for subsequent local training and trigger detection. Overall, the GSFS allows each client to upload \emph{asynchronously} after detecting significant changes (\ref{eq:perf-trigger}), but the server consolidates those asynchronous updates into a \emph{single broadcast} event. Regardless of how frequently or infrequently a client uploads, it always receives the newest global model at each broadcast iteration. This design reduces communication overhead compared to forcing uniform uploads from all clients after every local epoch.\\
\textbf{Cloud Aggregation:} The Federated Learning Coordinator aggregates updates from edge devices using privacy-preserving algorithms such as Federated Averaging. The global model leverages the high computational power of the cloud for large-scale updates, ensuring scalability and robustness. Communication between the central server (or aggregator) and edge clients is a critical component of federated learning frameworks. It must handle model updates, client feedback, and synchronization signals in a robust and secure manner. In our system, we use TCP/IP-based Transport, which is known as a widely adopted mechanism due to its simplicity and compatibility with most platforms. The central server and clients exchange model parameters over standard TCP sockets, secured by TLS/SSL, to ensure data confidentiality and integrity.

\subsubsection{Datasets}

The training uses dataset IoT-23 (network traffic data) to improve model performance in IoT-specific tasks. These datasets are preprocessed to align with the LLM’s input requirements and enhance the model’s capabilities in predictive analytics and resource management.

\subsubsection{Experimental Setup}

To evaluate the proposed framework, we conducted experiments on a two-tier platform. The first tier, referred to as the central infrastructure, provided high-performance resources for training and serving LLMs. Specifically, we used an Ubuntu 22.04 operating system running on an AMD EPYC server equipped with an NVIDIA A100 (40\, GB) GPU. The second tier, denoted as the client environment, was deliberately limited in both memory and GPU capacity compared to central infrastructure to reflect realistic edge or IoT scenarios. We utilized a system with 16\, GB system memory and a single NVIDIA GeForce RTX 3080\ Ti GPU featuring 12\, GB of VRAM. Under these constraints, we deployed smaller or distilled versions of the trained models to demonstrate adaptability and latency performance in resource-limited settings. This setup allowed us to assess how well the framework adapts when facing practical deployment conditions that cannot accommodate large-scale models. To evaluate the framework, we used a 
experimental setup comprising heterogeneous IoT devices and a cloud infrastructure. The dataset IoT-23 was preprocessed and used for training the LLMs. The evaluation focused on metrics such as accuracy, recall, precision, response latency, energy efficiency, etc. 

\section{Results and Discussion}
\label{sec:Results}

\begin{table}[h]
\centering
\caption{Dataset Split Sets}
\label{tab:dataset_split}
\begin{tabular}{lcc}
\toprule
\textbf{Dataset Part} & \textbf{Percentage (\%)} & \textbf{Number of Samples} \\ 
\midrule
Training Set          & 70                      & 33,602                     \\ 
Validation Set        & 20                      & 9,601                      \\ 
Testing Set           & 10                      & 4,800                      \\ 
\bottomrule
\end{tabular}
\end{table}

\begin{table*}[ht]
\centering
\caption{Performance Comparison of Different Models: 
Each row reports precision, recall, F1-score, and accuracy for classification performance, 
along with response latency (in milliseconds) and energy efficiency (requests per minute).}
\label{tab:performance_comparison}
\begin{tabular}{lcccccc}
\toprule
\textbf{Models} & \textbf{Precision} & \textbf{Recall} & \textbf{F1} & \textbf{Accuracy} & \textbf{Response Latency (ms)} & \textbf{Energy Efficiency (req/min)} \\ 
\midrule
BERT                     & 0.9058 & 0.8932 & 0.8741 & 0.8932 & 115.48 & 519.40 \\
ALBERT                   & 0.9073 & 0.9001 & 0.8846 & 0.9015 & 110.62 & 542.50 \\
PALM                     & 0.8965 & 0.8917 & \textbf{0.8939} & 0.8921 & 165.83 & 361.90 \\
ROBERTA                  & 0.8949 & 0.8905 & 0.8625 & 0.8710 & 145.91 & 411.30 \\
DEBERTA                  & 0.9053 & 0.9012 & 0.8931 & 0.8824 & 140.48 & 427.20 \\
XLNET                    & 0.9062 & 0.8923 & 0.8743 & 0.8935 & 130.67 & 459.30 \\
GPT-Neo                  & 0.9017 & 0.8920 & 0.8730 & 0.8920 & 175.34 & 342.30 \\
GPT-2 (124M)             & 0.9005 & 0.8912 & 0.8721 & 0.8910 & 125.54 & 478.00 \\
GPT-2-Medium (355M)      & 0.9025 & 0.8930 & 0.8742 & 0.8931 & 195.62 & 306.70 \\
GPT-2-Large (774M)       & 0.9012 & 0.8914 & 0.8726 & 0.8922 & 330.53 & 181.50 \\
google/flan-t5-base      & 0.7132 & 0.6124 & 0.5904 & 0.7019 &  95.48 & 628.60 \\
google/flan-t5-large     & 0.7346 & 0.6213 & 0.6033 & 0.7066 & 165.96 & 361.50 \\
EleutherAI/pythia-410M   & 0.9045 & 0.8913 & 0.8757 & 0.8969 & 145.42 & 412.60 \\
EleutherAI/pythia-1b     & 0.9111 & 0.9001 & 0.8813 & 0.9008 & 280.62 & 213.80 \\
bigscience/bloomz-560m   & 0.8645 & 0.8412 & 0.8438 & 0.8662 & 170.48 & 352.00 \\
bigscience/bloomz-1b1    & 0.8713 & 0.8561 & 0.8510 & 0.8705 & 310.84 & 193.00 \\
OPT-350M                 & 0.9065 & 0.8904 & 0.8793 & 0.8971 &  \textbf{88.14} & \textbf{680.70} \\
OPT-1.3b                 & \textbf{0.9172} & \textbf{0.9083} & 0.8878 & \textbf{0.9054} & 215.36 & 278.70 \\
\bottomrule
\end{tabular}
\end{table*}

The proposed framework demonstrated significant improvements in IoT management tasks. Edge processing reduced latency by enabling real-time decision-making, while federated learning ensured data privacy and scalability. The adaptive feedback loop enhanced the system’s ability to adjust to dynamic IoT environments by leveraging continuous performance monitoring and feedback. The monitoring dashboard provided administrators with actionable insights, enabling quick adjustments to optimize system operations. Additionally, the distributed architecture enhanced resilience to failures and demonstrated scalability as the number of devices increased. Energy efficiency was significantly improved due to localized processing on edge devices, reducing reliance on cloud resources.

\begin{figure}[ht!]
    \centering
    \includegraphics[width=8.7cm]{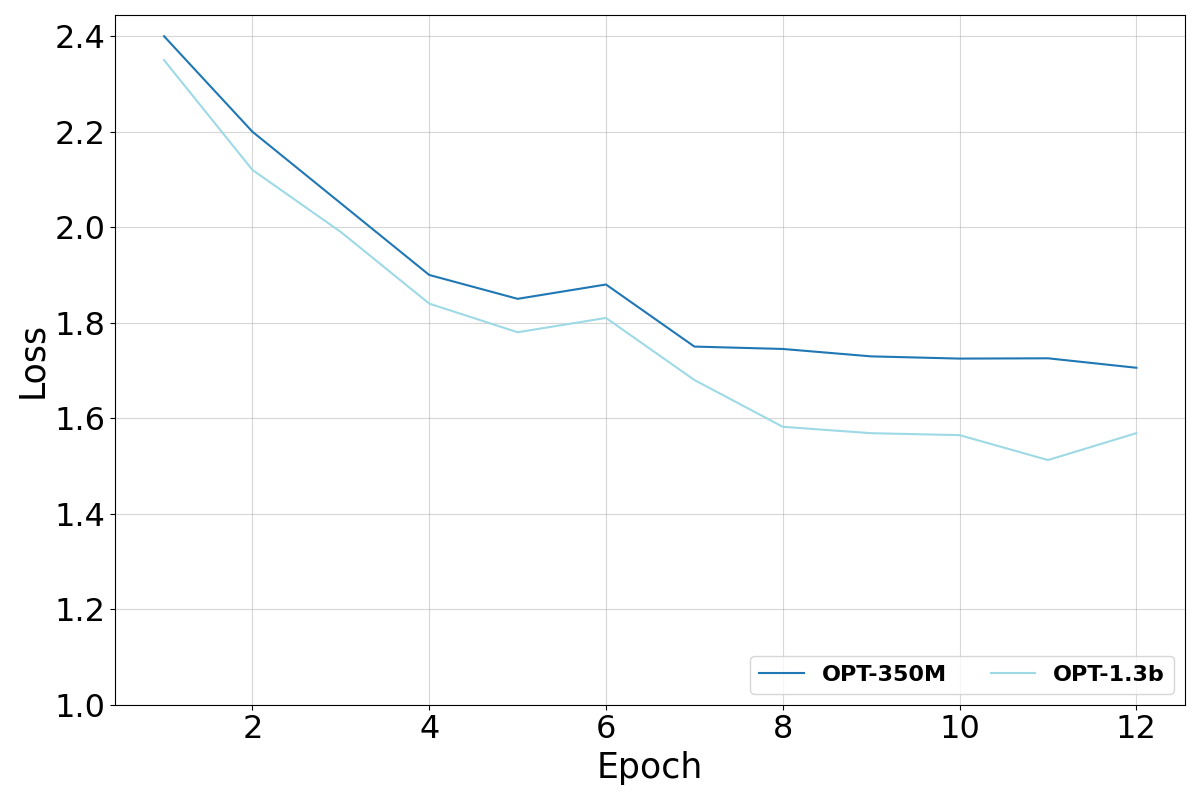}
    \caption{Loss curves for the OPT-350M and OPT-1.3b models over training epochs.}
    \label{fig:loss_curves}
\end{figure}

\subsection{Model Performance Results}

\begin{table*}[ht!]
\centering
\caption{
Comparison of FedAvg, FedOpt and the proposed GSFS approach in terms of performance and efficiency.
}
\label{tab:fl_strategy_comparison}
\begin{tabular}{lcccc}
\toprule
\textbf{Strategy} & \multicolumn{2}{c}{\textbf{Accuracy}} & \multicolumn{2}{c}{\textbf{F1-score}} \\
\cmidrule(lr){2-3} \cmidrule(lr){4-5}
 & \textbf{Central Model} & \textbf{Client Model (avg)}
 & \textbf{Central Model} & \textbf{Client Model (avg)} \\
\midrule
FedAvg & 0.8945 & 0.8732 & 0.8761 & 0.8619 \\
FedOpt & 0.8982 & 0.8522 & 0.8641 & 0.8685 \\
GSFS (Proposed) & 0.9008 & 0.8881 & 0.8825 & 0.8754 \\
\bottomrule
\end{tabular}
\bigskip
\begin{tabular}{lcccc}
\toprule
\textbf{Strategy} & \multicolumn{2}{c}{\textbf{Latency (s)}} & \multicolumn{2}{c}{\textbf{Energy Efficiency (Req/min)}} \\
\cmidrule(lr){2-3} \cmidrule(lr){4-5}
 & \textbf{Central Model} & \textbf{Client Model (avg)}
 & \textbf{Central Model} & \textbf{Client Model (avg)} \\
\midrule
FedAvg      & 48.12 & 71.43 & 1.25 & 0.84 \\
FedOpt     & 42.09 & 55.08 & 1.43 & 1.09 \\
GSFS (Proposed) & 37.74 & 34.69 & 1.59 & 1.73 \\
\bottomrule
\end{tabular}
\end{table*}

We separated the dataset into three parts: 70\% for training, 20\% for validation, and 10\% for testing. The dataset we used is the cleaned version of the IoT-23 dataset. For the original dataset, rows with missing data or encoding errors were removed during the cleaning process. After cleaning, we obtained a dataset containing a total of 48,003 samples. Table \ref{tab:dataset_split} presents the dataset partitioning into training, validation, and testing subsets, including the number of samples in each subset. This section also presents the quantitative results and evaluation curves for the base LLM model selection. To be more specific, two of the performance metrics were defined as follows: \textbf{Response Latency}: The average inference time required for the model to process the test set.
\textbf{Energy Efficiency}: The number of requests the model can handle within one minute. Table~\ref{tab:performance_comparison} presents the performance of various LLMs on the cleaned IoT-23 dataset, evaluating the accuracy, precision, recall, F1-score, response latency, and energy efficiency. Models with over 1B parameters were trained as central models on an A100 GPU, while smaller models operated as client models on an RTX 3080Ti GPU. Bert-based models were included for comparison and trained on the RTX 3080Ti due to their smaller sizes. The results indicate that large-scale LLMs generally achieve higher classification metrics but at the expense of increased latency and reduced energy efficiency. ALBERT strikes a balance, delivering a strong F1-score with moderate latency, whereas some GPT-2 variants exhibit slower inference times. Notably, OPT-1.3B achieves the highest precision (0.9172) and recall (0.9083), with an accuracy of 0.9054. In terms of response latency and energy efficiency, larger models tend to perform better in classification but suffer from increased computational costs. OPT-350M demonstrates the lowest response latency (88.14 ms) while maintaining competitive classification metrics. Meanwhile, T5-based models, such as \texttt{google/flan-t5-base} and \texttt{google/flan-t5-large}, show relatively lower classification scores but offer competitive latency and higher throughput under specific conditions. The loss curves in Figure \ref{fig:loss_curves} further support the selection of the OPT model, showing a consistent downward trend, with loss decreasing rapidly in the early epochs and gradually stabilizing in later epochs. A small upward spike is observed around Epochs 6 and 7, possibly due to momentary fluctuations during optimization. This stable training behaviour indicates that the OPT model is a reliable choice for further experiments. Considering the performance across multiple metrics and their faster inference speeds, we chose OPT series models for subsequent experiments in this study.

\subsection{Federated Learning Strategy's Comparison}

The performance of the central and client models using the proposed GSFS is evaluated against the classical FedAvg and FedOpt approaches. Table~\ref{tab:fl_strategy_comparison} presents key evaluation metrics, including Accuracy, F1-score, Response Latency, and Energy Efficiency, after 10 rounds of federated training. The Central Model column represents the aggregated model at the server, while the Client Model (avg) column reports the averaged performance across participating clients. In terms of latency, FedAvg exhibits the highest delay, as it relies on full-parameter aggregation at the central server after each round of training. In contrast, FedOpt reduces latency by optimizing communication and update efficiency. GSFS further minimizes response time by employing a partial parameter update mechanism, selectively transmitting model updates based on significant local changes. A similar trend is observed in energy efficiency, measured as the number of requests processed per minute. GSFS achieves higher efficiency than FedAvg by reducing redundant updates, optimizing client-server synchronization, and balancing computational load across nodes. Overall, the results indicate that GSFS consistently outperforms FedAvg in both central and client models, delivering superior Accuracy and F1-score while significantly improving latency and energy efficiency. These findings demonstrate the effectiveness of the proposed strategy in federated learning for IoT environments.

\section{Conclusion}
\label{sec:Conclusion}

This paper presents an adaptive IoT management framework that integrates LLMs with federated learning to enhance scalability, intelligence, and data privacy in distributed IoT environments. By leveraging an adaptive federated updating strategy, the framework dynamically optimizes client participation and learning rates, reducing communication overhead while maintaining model accuracy. Additionally, the hybrid edge-cloud architecture ensures a balanced computational workload, enabling real-time decision-making at the edge while utilizing cloud resources for global model refinement. The experimental results, conducted using the IoT-23 dataset, demonstrate that the proposed approach improves malware detection and anomaly detection in IoT networks. Compared to traditional federated learning methods like Fed\_Avg and Fed\_Opt, our framework achieves higher model accuracy, lower response latency, and enhanced energy efficiency. The results validate the effectiveness of combining LLMs with federated learning for intelligent IoT management. 

\end{document}